\documentclass[letterpaper]{article} 
\usepackage{aaai25}  
\usepackage{times}  
\usepackage{helvet}  
\usepackage{courier}  
\usepackage[hyphens]{url}  
\usepackage{graphicx} 
\usepackage{natbib}  
\usepackage{caption} 
\usepackage{algorithm}
\usepackage{algorithmic}
\usepackage{amsmath}
\usepackage{longtable,tabularx}
\usepackage{threeparttable}
\usepackage{booktabs}
\usepackage{subfig}
\usepackage{siunitx}

\usepackage{newfloat}
\usepackage{listings}
\DeclareCaptionStyle{ruled}{labelfont=normalfont,labelsep=colon,strut=off} 
\floatstyle{ruled}
\newfloat{listing}{tb}{lst}{}
\floatname{listing}{Listing}
\title{Semi-Markovian Planning to Coordinate Aerial and Maritime \\ Medical Evacuation Platforms}
\author {
    Mahdi Al-Husseini\textsuperscript{\rm 1},
    Kyle H. Wray\textsuperscript{\rm 2},
    Mykel J. Kochenderfer\textsuperscript{\rm 1}
}
\affiliations {
    \textsuperscript{\rm 1}Stanford University, 
     \textsuperscript{\rm 2} University of Massachusetts Amherst\\
    mah9@stanford.edu, kwray@umass.edu, mykel@stanford.edu
}

\usepackage{bibentry}

\begin{document}

\maketitle

\begin{abstract}
The transfer of patients between two aircraft using an underway watercraft increases medical evacuation reach and flexibility in maritime environments. The selection of any one of multiple underway watercraft for patient exchange is complicated by participating aircraft utilization history and a participating watercraft position and velocity. The selection problem is modeled as a semi-Markov decision process with an action space including both fixed land and moving watercraft exchange points. Monte Carlo tree search with root parallelization is used to select optimal exchange points and determine aircraft dispatch times. Model parameters are varied in simulation to identify representative scenarios where watercraft exchange points reduce incident response times. We find that an optimal policy with watercraft exchange points outperforms an optimal policy without watercraft exchange points and a greedy policy by 35\% and 40\%, respectively. In partnership with the United States Army, we deploy for the first time the watercraft exchange point by executing a mock patient transfer with a manikin between two HH-60M medical evacuation helicopters and an underway Army Logistic Support Vessel south of the Hawaiian island of Oahu. Both helicopters were dispatched in accordance with our optimized decision strategy.
\end{abstract}

\section{Introduction}

\begin{figure}[t]
    \centering
    \subfloat[\centering ]{{\includegraphics[width=0.52 \columnwidth]{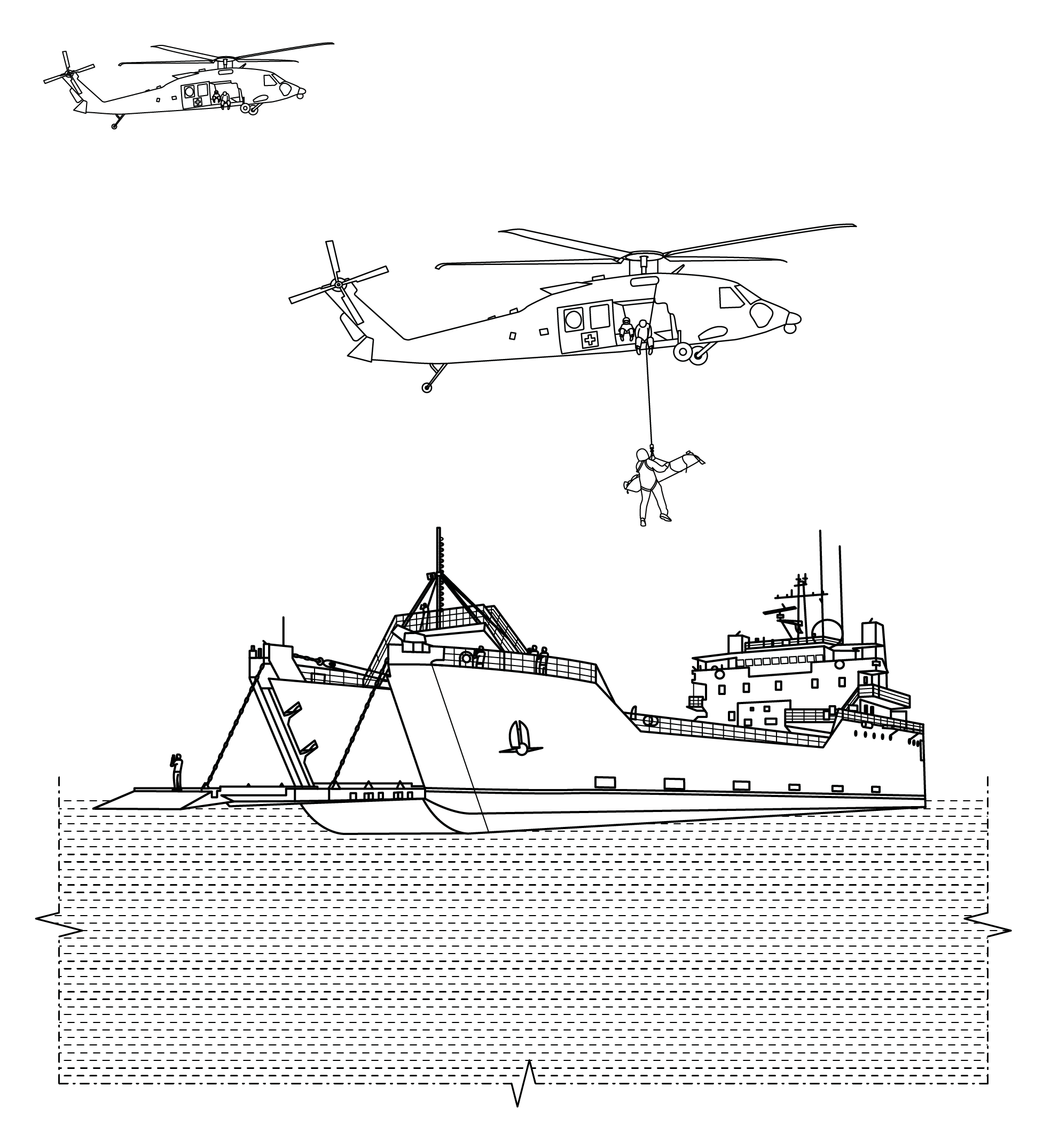} }}%
    \subfloat[\centering ][$\copyright$ Mahdi Al-Husseini 2023]{{\includegraphics[width=0.45 \columnwidth]{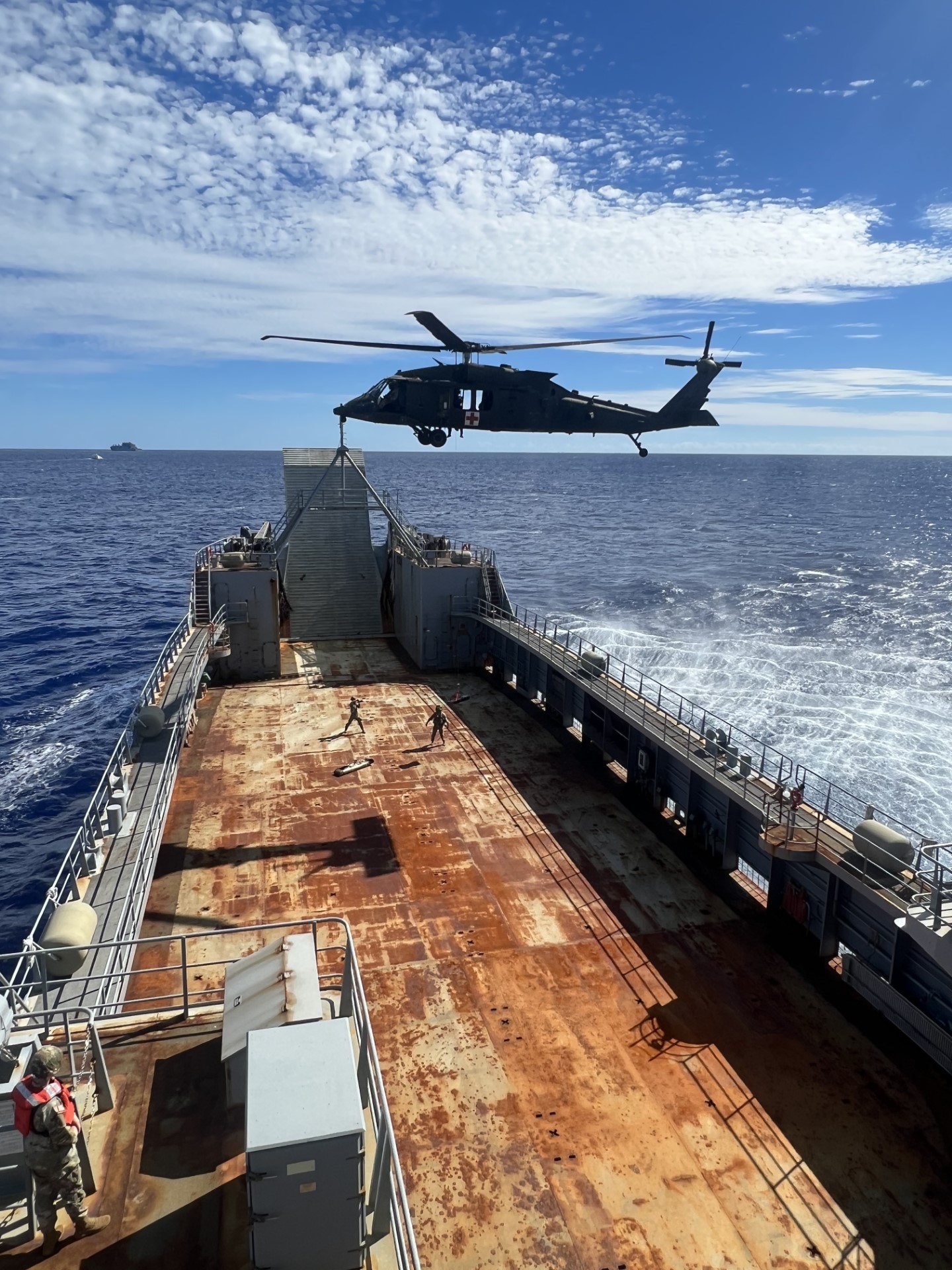} }}%
    \caption{An evacuation aircraft lowers a patient onto a watercraft while a second aircraft circles nearby. Using watercraft as exchange points between aircraft expedites patient movement and enables evacuation across vast distances.}%
\end{figure}



Medical planners coordinate multiple aerial and maritime evacuation platforms to facilitate the transfer of patients across expansive, non-contiguous maritime environments. To do so effectively involves understanding each platform's patient carrying capacity, position, forward speed, and other transportation and usage constraints, as well as the operational environment in time and space, e.g. casualty estimate data. The trajectory of watercraft, often underway in support of non-medical mission requirements, further complicates the exchange point selection process. Due to the challenging nature of the multi-agent coordination problem, patients in maritime environments are currently transferred via evacuation aircraft or hospital ships, or occasionally, from evacuation aircraft to hospital ships. Transportation exclusively by evacuation aircraft is fast, but transport distance is limited by aircraft range. Transportation exclusively by hospital ship can support any transport distance, but is slow. This paper considers, for the first time the use of underway watercraft as intermediary exchange points for two medical evacuation aircraft arriving from different islands (Fig. 1). Doing so combines the strengths of both aerial and maritime evacuation platforms, but requires careful selection and dispatching to minimize transport delays. This novel capability is inspired by the ambulance exchange point, which uses fixed, land-based points for patient exchange between ambulances \cite{medevac2019}. Unlike traditional ambulance exchange points, however, watercraft exchange locations are neither preidentified nor fixed in place. 

The use of watercraft as exchange points raises questions: 
\begin{itemize}
  \item Which watercraft should be selected as an exchange point to minimize incident response time for a given evacuation request? 
  \item How does a given watercraft exchange point affect a participating aircraft's ability to support future missions?
  \item How does aircraft airspeed, total casualty magnitude, distribution of patients between islands, expected number of patients per evacuation request, and severity of patient injury affect exchange point selections? 
\end{itemize}


To enable this capability, we first model a representative dynamic environment with multiple aircraft and watercraft operating across and between two islands, then introduce a sequential decision making process and online solver to select optimal watercraft exchange points and determine aircraft dispatching. Our paper's main contributions are:

\begin{itemize} 
  \item Develop a model environment and semi-Markov decision process (SMDP) \cite{Hu2008,baykal2007semi} for the coupled watercraft exchange point selection and aircraft dispatch problems. 
  \item Introduce an online planner that applies Monte Carlo Tree Search (MCTS) with root parallelization \cite{chaslot2008parallel} using patient evacuation request chains from a casualty generation model.
  \item Adjust five model parameters to identify representative scenarios where watercraft exchange points would expedite transfers. Cruise speeds of aircraft with historic, current, or proposed evacuation roles are considered. 
  \item Deploy the watercraft exchange point in Hawaii using two HH-60M Black Hawk helicopters and an Army Logistics Support Vessel (LSV). The dispatch of helicopters to transfer a mock patient was informed by our model, resulting in minimal delay during patient exchange. 
\end{itemize} 

\section{Related Work}

To our best knowledge, this is first effort to apply semi-Markovian planning to the multi-platform maritime evacuation coordination problem. It is also the first to develop a planning framework for the proposed watercraft exchange point capability, the underlying mechanisms of which we explore in a prior companion paper \cite{watercraftaxp}. There, we provide the problem description and operational considerations for air medical experts. In contrast, our research presented here details the artificial intelligence, computation, and modeling contributions as well as experimental results. The literature introduces adjacent efforts for medical evacuation resource allocation, characteristics of which feature prominently in our work. 



\citet{mclay2013dispatching} develop a linear programming method for dispatch, and consider several equity measures to ensure an equalized distribution of resources across various demand signals. \citet{keneally2016markov} develop an MDP that dispatches medical evacuation aircraft in a land-based combat environment over an extended period of time to maximize overall system utility. \citet{jenkins2018examining} build on efforts by \citeauthor{keneally2016markov}, introducing a monotonically decreasing function over time to more accurately reflect patient survival probability. \citet{pettet2021hierarchical} present a hierarchical framework for partitioning large evacuation dispatch and allocation problems into tractable intertwined sub-problems. To overcome the sparsity of evacuation incidents available for consideration, they generate several casualty threads using an internal prediction model and instantiate individual MCTS trees for each. Equity considerations, a monotonically decreasing reward function, and generated casualty threads that align with parallel search trees all feature in this paper. 




\section{Problem Formulation}

\subsection{Model Environment}

\begin{figure*}[t]
\centering
\includegraphics[width=0.99\textwidth]{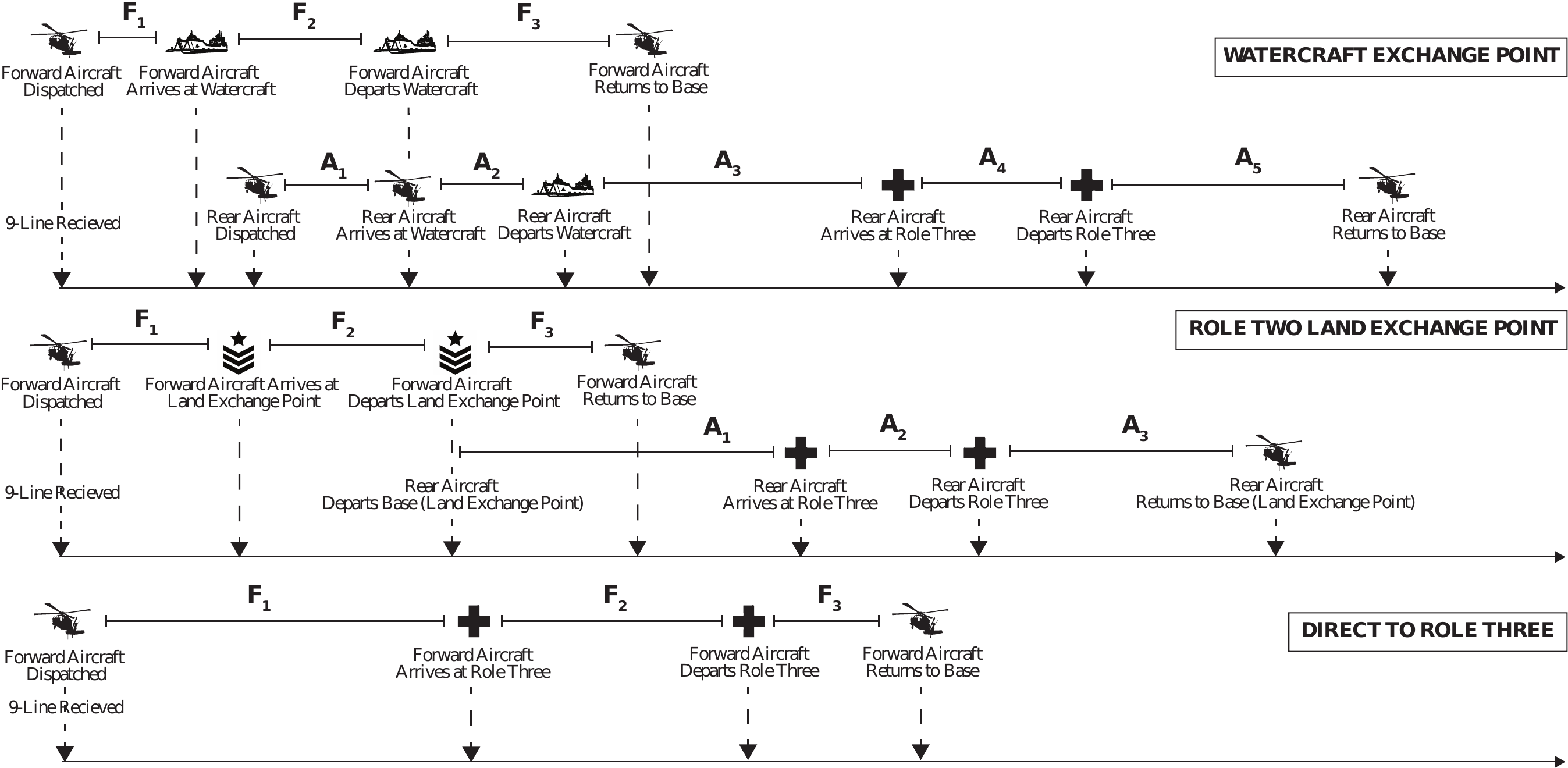}
\caption{Action sequences for three categories of actions: watercraft exchange points, land exchange point, and direct to role three transfers. Time spans $F_i$ are associated with the action sequence for the Kauai aircraft, while time spans $A_i$ are associated with the action sequence for Oahu aircraft. $F_1$ includes the time required to pickup the patient from point of injury.}
\end{figure*}

The model scenario involves a large scale combat operation across the Hawaiian Islands of Oahu and Kauai. Patients are induced sequentially in accordance with a Poisson point process. Each patient is evacuated to the next higher role of care in accordance with Army doctrine \cite{health2020}. Evacuation requests flow from a role one (first responder) to a role two (forward resuscitative care) to a role three (theater hospitalization) \cite{cunningham2019military}. Patients begin at one of eight spatially distributed role one or role two facilities. Medical evacuation platoons with HH-60M Black Hawk helicopters are staged on both islands at role twos. We refer to the forward support medical evacuation platoon (FSMP) aircraft as the forward or Kauai aircraft, and the area support medical evacuation platoon (ASMP) aircraft as the rear or Oahu aircraft. FSMP aircraft are postured to coincide with demand, typically near troops engaged in combat or in high population density areas. ASMP aircraft move patients between treatment facilities on an area basis. Three military watercraft, an LSV, a Landing Craft Utility (LCU), and an Expeditionary Fast Transport (EPF) travel between both islands at \SI{10}{knots}, \SI{8}{knots}, and \SI{43}{knots} respectively. As described in Fig. 2, transfers may employ watercraft exchange points, land exchange points, or may fly direct to the role three. The distance between Oahu and Kauai is such that a typical rotary-wing aircraft can traverse it one way without requiring an intermediary transfer or refuel point, allowing fixed land and moving watercraft exchange points to be considered simultaneously. Each received evacuation request contains the originating location, the number of patients to be transported, and their transport destination. On evacuation request receipt, the appropriate aircraft is dispatched to the patient pickup site. After collection, the aircraft delivers its patients to either an exchange point or the transport destination. The aircraft then returns to base where it refuels and subsequently enters an idle state. Please see \citet{watercraftaxp} for additional problem description details. 



\subsection{Semi-Markov Decision Process Formulation}
We introduce a fully-observable semi-Markov decision process (SMDP) to model the environment and aid in exchange point selection. The SMDP is represented by $\mathcal{M} = \langle \mathcal{S, A}, P, \gamma, R, \Upsilon \rangle$ where $\mathcal{S}$ is the state space, $\mathcal{A}$ is the action space, $P$ is the transition model, $\gamma$ is the discount factor, $R$ is the reward function, and $\Upsilon$ is the sojourn model \cite{baykal2010semi}. Each state $s$ in $\mathcal{S}$ includes the mission delay for participating aircraft, the likelihood a participating aircraft will have a maintenance fault and require replacement, the historic utilization rate of participating aircraft, and the considered evacuation request. The action space $\mathcal{A}$ includes the LSV exchange point, LCU exchange point, EPF exchange point, Wheeler Army Airfield exchange point, and direct to Tripler exchange point. The transition model $P$ assigns a probability of transitioning from an existing state $s$ by action $a$ to state $s'$, with $\sum_{s \in \mathcal{S}} P(s, a, s') = 1$. The discount factor $\gamma$ is a hyperparameter that balances short and long-term rewards. A larger $\gamma$ prioritizes long-term rewards, whereas a smaller $\gamma$ prioritizes short-term rewards. Sojourn model $\Upsilon$ represents the distribution over sojourn time condition on current state $s$ and next state $s'$. The sojourn time is the state transition duration, or the time required to resolve an evacuation request. The SMDP is an appropriate model for the exchange point selection problem, because the sojourn time duration for a given state transition has a significant impact on accumulated rewards, given the presence of intermediary evacuation requests. 


We develop separate reward functions for greedy and optimized policies, the latter which includes intermediate rewards for point of injury evacuations between decision epochs. Both greedy and optimized policies employ a monotonically decreasing fusion of a Weibull survival function with a linear reward function, as shown in Fig. 3. A survival function is the complimentary cumulative distribution over a lifetime, and it represents the probability a casualty will survive after a certain amount of time has elapsed following injury. The Weibull survival function extends an exponential distribution to the hazard rate \cite{zhang2011research}. This represents the non-linear nature of the physiological ``golden hour'' employed in military medical evacuation asset allocation policy. The golden hour reflects the belief that trauma patients are far more likely to survive if definitive care is provided within sixty minutes of sustaining an injury \cite{kotwal2016effect}. Evacuation operations during large scale combat must also consider battlefield clearance requirements \cite{hamiltonclearing}, here reflected in the linear component of our reward function. 



\begin{figure}[t]
\centering
\includegraphics[width=0.99\columnwidth]{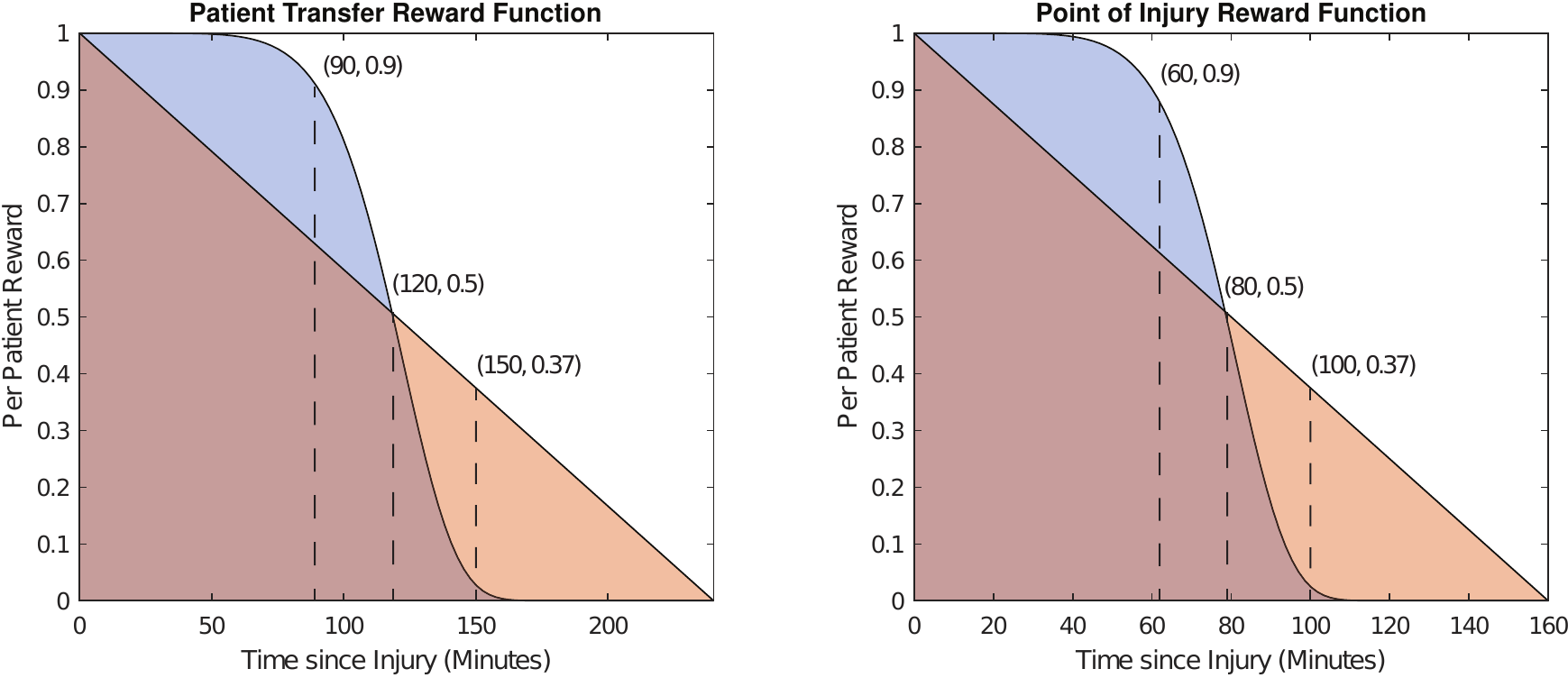}
\caption{Fused reward functions capturing non-linear survivability estimates in blue and linear battlefield clearance requirements in red.}
\end{figure}


The generic per patient fused reward function $R_{f}$ formulation follows, where $t$ is time since injury, $m$ is the battlefield clearance slope, and $a$ and $\gamma$ are Weibull cumulative distribution function parameters:
\small
\begin{equation}
  R_f = \max{(e^{-(\frac{t}{a})^\gamma}, -mt+1)}
\end{equation}
\normalsize

The greedy policy reward function is therefore the per patient fused reward function times the number of patients transported $P$:
\small
\begin{equation}
  R_{\text{greedy}} = R_f(a=125, \gamma=7, m=0.0042, t) \cdot P
\end{equation}
\normalsize

$R_f$ applied with the above values for $a$, $\gamma$, and $m$ indicates that a patient transfer completed within 90 minutes of lifesaving stabilization at a role two has a 90\% probability of survival, whereas a patient transfer completed 120 minutes of lifesaving stabilization has a 50\% probability of survival. The fusion reward function may be easily adjusted for injury severity and type and means of stabilization at the role two, and per command-driven requirements for battlefield clearance during different phases of a military operation. The reward function for optimized policies $R_{\text{optimal}}$ adds intermediate rewards to patient transfer rewards. Intermediate rewards are incurred from and summed across point of injury evacuations that occur between patient transfer decision epochs. Intermediate rewards capture the negative impact patient transfer delays have on subsequent point of injury evacuations. The optimal per patient fused reward function $R_{\text{optimal}}$ formulation follows, where $x$ is a point of injury evacuation request in the set of requests $\mathcal{X}$ which occur between decision epochs:

\small
\begin{equation}
\begin{aligned}
  R_{\text{optimal}} & =  \sum_{x \in \mathcal{X}} R_f(a=63, \gamma=7, m=0.0063, t_x) \cdot P_x \\
                     &   + R_f(a=125, \gamma=7, m=0.0042, t)
\end{aligned}
\end{equation}
\normalsize

Time since injury $t$ is equal to incident response time $T$ plus time between the injury occurring and the evacuation asset being dispatched. The former is a function of the exchange point selected and calculated in accordance with the action sequences in Fig. 2, while the latter depends on delays from queuing of requests to assets. $T$ is calculated as follows, where $F$ is a time block associated with movement of the forward aircraft, $A$ is a time block associated with movement of the rear aircraft, and exchange point $E$ may be in the set of watercraft exchange points $W$ or land role two exchange points $L$:

\small
\begin{equation}
  T =
    \begin{cases}
      [F_1 + F_2 + A_2 + A_3] & \text{if } E \in W\\
      [F_1 + F_2 + A_1 + A_2] & \text{if } E \in L\\
      [F_1] & \text{otherwise}\\
    \end{cases}
\end{equation}
\normalsize

Intermediate rewards may be supplemented with a penalty $Y$. $Y$ penalizes each aircraft $h$ in the set of participating aircraft $H$ for time of employment $T_{h}$ relative to historic aircraft utilization $U_h$. $T_{h}$, like $T$, is a consequence of exchange point selection. The penalty formulation follows, where $\tau_1$ and $\tau_2$ are tunable parameters: 

\small
\begin{equation}
  T_{h_1} = [F_1 + F_2 + F_3]
\end{equation}
\normalsize

\small
\begin{equation}
  T_{h_2} =
    \begin{cases}
      [A_1 + A_2 + A_3 + A_4 + A_5] & \text{if } E \in W\\
      [A_1 + A_2 + A_3] & \text{if } E \in L\\
      [0] & \text{otherwise}\\
    \end{cases}
\end{equation}

\small
\begin{equation}
  Y = \tau_1 T_{h1} U_{1} + \tau_2 T_{h2} U_{2}
\end{equation}
\normalsize


\subsection{Monte Carlo Tree Search}


An expansive military campaign across a non-contiguous maritime environment with multiple heterogeneous evacuation platforms results in complex patient evacuation dynamics that prove difficult to model in closed form. These evacuation models motivate the use of probabilistic search algorithms, such as MCTS, in tandem with a generative environment. MCTS is an anytime algorithm, meaning it can return a valid, if sub-optimal, solution at any point in its run-time while also adapting to changes in the model environment \cite{browne2012survey}. MCTS represents the set of possible actions as edges in a search tree. The nodes in the search tree are then the ensuing states. MCTS expands the search tree using random sampling or a predefined heuristic informed by the problem domain. MCTS explores the search tree asymmetrically such that the most promising actions are prioritized. State value estimation occurs by simulating a play-out from a node to the end of a predefined planning horizon. We apply the standard Upper Confidence bound for Trees (UCT) algorithm \cite{kocsis2006improved} to balance exploration versus exploitation in the tree search policy and determine the value of each node $n$, as follows:

\small
\begin{equation}
  \text{UCT}(n) = \bar{u}(n) + c \sqrt{\frac{\text{log(visits}(n)}{\text{visits}(n')}}
\end{equation}
\normalsize
\noindent
such that $\bar{u}(n)$ is the value of the state at $n$, $c$ is an exploration constant that adjusts the balance between exploration of new nodes and exploitation of previously visited nodes, visits($n$) is the visitation count for $n$, and visits($n'$) is the visitation count for the parent node of $n$, $n'$. A default, heuristic policy is typically applied to estimate action value during search tree roll-outs. We introduce a greedy policy for roll-outs that selects the exchange point which, based on proximity alone and without accurate state information, minimizes transfer time for the evacuation request under consideration. 

\subsection{Root Parallelization}

Root parallelization is a process whereby multiple independent search trees are constructed and  solved using MCTS \cite{chaslot2008parallel}. The search trees differ in composition due to stochastic sampling during construction. Results are then aggregated across all search trees to select an optimal course of action. In this paper, search trees are aligned to evacuation request ``threads'' generated from simulated casualty data. Search tree ``scores'' may be combined across search trees that begin with a particular action, in which case the action with the greatest summed score is selected. Sampling a single evacuation request thread proves insufficient, due to significant environmental uncertainty. Root parallelization paired with casualty generation data can provide robust results even when actual evacuation requests in theater are sparse.


\subsection{Parameters}
Environmental parameters can be grouped into evacuation support and evacuation request parameters. We introduce two evacuation support parameters: exchange point selection action space and dispatch policy. SMDP effectiveness is evaluated for action spaces $\mathcal{A}_1$ and $\mathcal{A}_2$. $\mathcal{A}_1$ includes all moving watercraft exchange points, fixed land exchange points, and the direct to role three transfer. $\mathcal{A}_2$ features only fixed land exchange points and the direct to role three transfer. We compare a greedy policy against an online solver in MCTS applied to the SMDP model. The main source of environmental uncertainty is the ebb and flow of patients across the battlefield, which we characterize using evacuation request parameters. These include casualty magnitude multiplier, platoon casualty ratio, proportion of patient transfers, and patients per evacuation request. Also considered is the cruise airspeed of the aircraft servicing the evacuation requests. 



\begin{table}[h!]
\fontsize{9pt}{10pt} \selectfont
\centering
\begin{tabular}{@{}l l@{}}\toprule 
 Parameter & Values \\
 \midrule 
 action space & [$\mathcal{A}_1$, $\mathcal{A}_2$] \\ 
 dispatch policy & [Greedy, SMDP with MCTS] \\ 
 casualty magnitude multiplier & [0.6, 0.7, \dots, 1.4] \\ 
 platoon casualty ratio & [0.6, 0.7, \dots, 1.4] \\
 proportion of patient transfers & [0.1, 0.2, 0.3, 0.4, 0.5] \\
 aircraft airspeed [kn] & [120, 130, 150, 250, 280] \\
 patients per evacuation request & [2, 3, 4, 5, 6] \\ 
 \bottomrule
\end{tabular}
\caption{Environmental parameters.}
\end{table}


We experiment with four hyperparameters for MCTS with root parallelization: thread count, thread duration, discount factor, and exploration constant. Considered values for each are shown in Table 2. A two-dimensional grid search compares thread count and duration against total reward, for a given optimal policy. Ten threads with a duration of ten hours each was selected for all simulations.


\begin{table}[h!]
\fontsize{9pt}{10pt} \selectfont
\centering
\begin{tabular}{@{}l l@{}}\toprule 
 Hyperparameter & Values \\
 \midrule 
 thread count & [4, 6, 8, 10, 12, 14, 16] \\ 
 thread duration & [\SI{4}{\hour}, \SI{6}{\hour}, \SI{8}{\hour}, \SI{10}{\hour}, \SI{12}{\hour}, \SI{14}{\hour}, \SI{16}{\hour}] \\ 
 discount factor & [0.90] \\
 exploration constant & [1.0] \\
 \bottomrule
\end{tabular}
\caption{MCTS with root parallelization hyperparameters.}
\end{table}

\vspace{-1em}

\section{Experiments}

The introduced watercraft exchange point model and planner are first evaluated in simulation, then deployed in a military exercise with an actual LSV and two HH-60M Black Hawks. This section covers the simulation results, deployment and deployment insights, and capability discussion.

The model environment was simulated 20 times for each configuration of Table 1 parameters to determine the impact watercraft exchange points have on total rewards and incident response times by platoon, as shown in Fig. 4. Considered is how each parameter affects the ratio of patient transfers conducted via watercraft versus either land exchange points or direct transfer. Unless otherwise stated, the casualty magnitude multiplier is 1.0, the platoon casualty ratio is 1.4, proportion of patient transfers is 0.25, aircraft speed is 150 knots, and patients per evacuation request is three. All ranges reflect a 95\% confidence interval.

\begin{figure*}[ht!]
\centering
\includegraphics[width=0.99\textwidth]{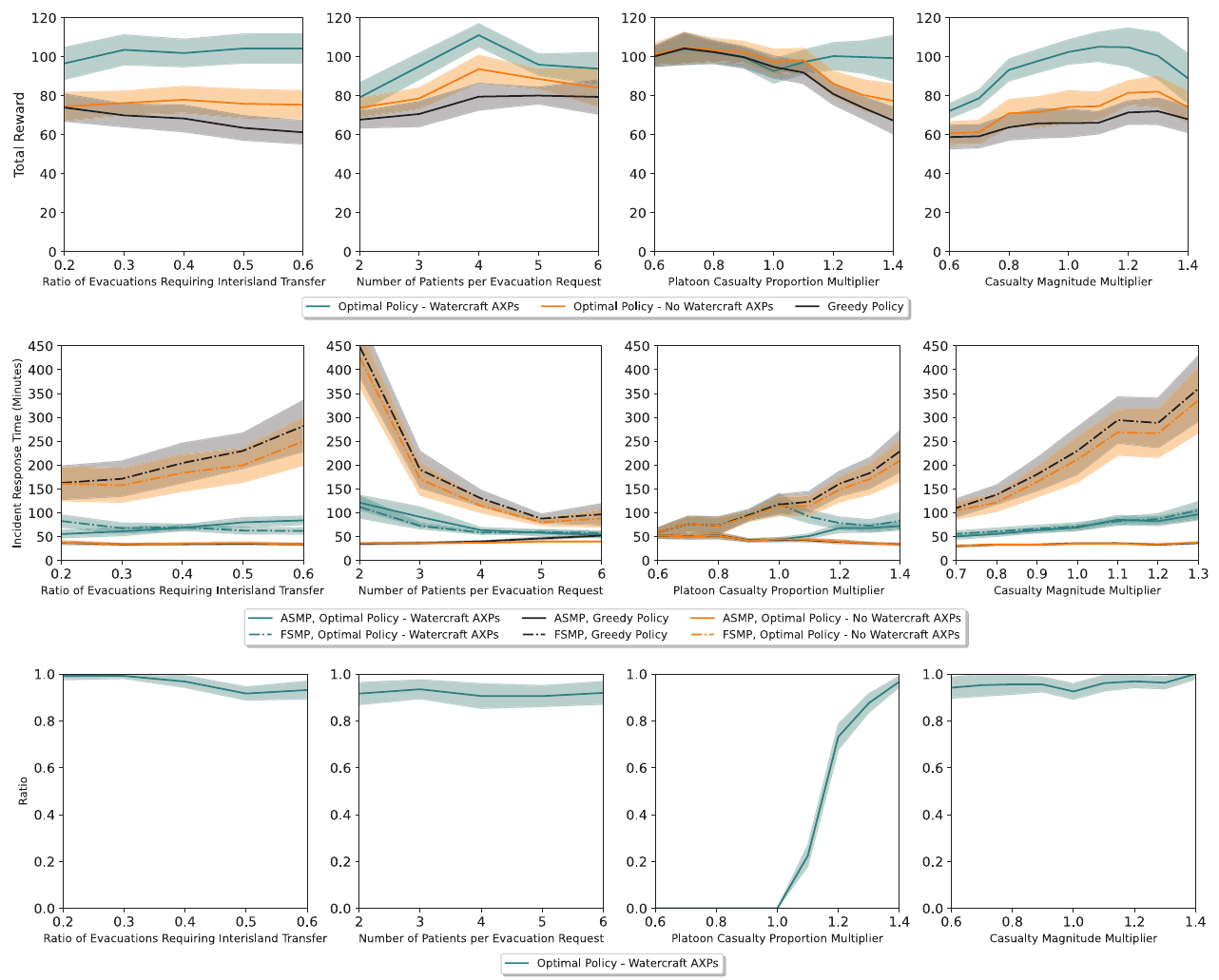}
\caption{Casualty magnitude, distribution of patients, patients per evacuation request, and proportion of interisland transfer patients impact on total rewards and incident response time by platoon for various exchange point policies, and ratio of patient transfers moved via watercraft given an optimal watercraft exchange point policy.}
\end{figure*}

\subsection{Total Casualty Magnitude}
Increasing the number of casualties increases total rewards. This increase is greater for an optimal policy with watercraft exchange points than for an optimal policy without watercraft exchange points. The evacuation system reaches a saturation point however, and total rewards begin to decrease across all three considered policies above a magnitude multiplier of 1.2. A casualty magnitude multiplier of 1.0 indicates 96 patients transferred via 32 evacuation requests distributed over 24 hours. Incidence response times increase dramatically with an increase in casualty magnitude for both an optimal policy without watercraft and a greedy policy. Incidence response times increased slightly with an increase in casualty magnitude for an optimal policy with watercraft. The ratio of patient transfers moved via watercraft does not change with an increase in casualty magnitude.


\subsection{Distribution of Patients Across Islands}


A geographic shift in casualty distribution significantly affects exchange point selection. Total reward and platoon incident response times remain steady if the majority of casualties are on Oahu and serviced by the ASMP aircraft. Watercraft exchange point availability increases total reward and reduces response times when the casualty magnitude proportion multiplier increases above 1.1. Watercraft exchange points are regularly employed by an optimal policy when the forward aircraft is sufficiently burdened relative to the rear aircraft. Between a 0.6 and 1.0 proportion multiplier, 0\% of patient transfers employed watercraft exchange points. By 1.4, more than 95\% of patient transfers employed watercraft exchange points. An optimal policy with watercraft exchange points outperforms both an optimal policy without watercraft exchange points and a greedy policy when the number of casualties on the forward island of Kauai exceed that on the rear island of Oahu by 25\% or more. 

\subsection{Number of Patients per Evacuation Request}
An increase in the number of patients transported per evacuation request, and a corresponding reduction in the number of total evacuation requests, increases total reward across optimal policies with and without watercraft exchange points. This is true for up to four patients, or two-thirds of an HH-60M's cabin capacity. The total reward for an optimal policy then begins to decrease for five or more patients per request. Watercraft exchange point availability results in a significant increase in total rewards when the number of patients per evacuation request averages three or four. The number of patients per evacuation request has minimal impact on the ratio of patient transfers moved via watercraft for an optimal policy with watercraft exchange points.

\subsection{Proportion of Interisland Transfer Patients}
A change in the proportion of interisland transfer patients has negligible impact on total rewards for both optimal policies with and without watercraft exchange points. There is a weak inverse relationship between the proportion of patients requiring interisland transfer and total rewards for a greedy policy. For an optimal policy with watercraft exchange points, an increase in the percentage of interisland transfers results in a decrease in FSMP response times and an increase in ASMP response times. The reverse is true for both an optimal policy without watercraft exchange points and a greedy policy. Changing the proportion of interisland transfer patients has little effect on use of watercraft exchange points. An optimal policy with watercraft exchange points significantly outperforms both an optimal policy without watercraft exchange points and a greedy policy across the full range of values considered. 

\subsection{Aircraft Airspeed}
Fig. 5 shows how watercraft exchange points affect incident response times across combinations of total casualty magnitude and aircraft cruise airspeeds. FSMP and ASMP aircraft are assumed to be homogeneous assets with the same cruise airspeeds. As discussed, watercraft exchange points reduce the dramatic spike in response times resulting from an increased number of casualties. The faster the aircraft cruise airspeed however, the less impact watercraft exchange points have on incident response times. Watercraft exchange points reduce average incident response times by 50\% for two Bell UH-1 Iroquois and a 30\% increase in total casualties. In comparison, the SB-1 Defiant and Bell V-280 Valor minimally reduce incident response times when there is a 30\% increase in casualties and watercraft exchange points are available (16\% and 14\% respectively).


\begin{figure}[ht]
\centering
\includegraphics[width=0.99\columnwidth]{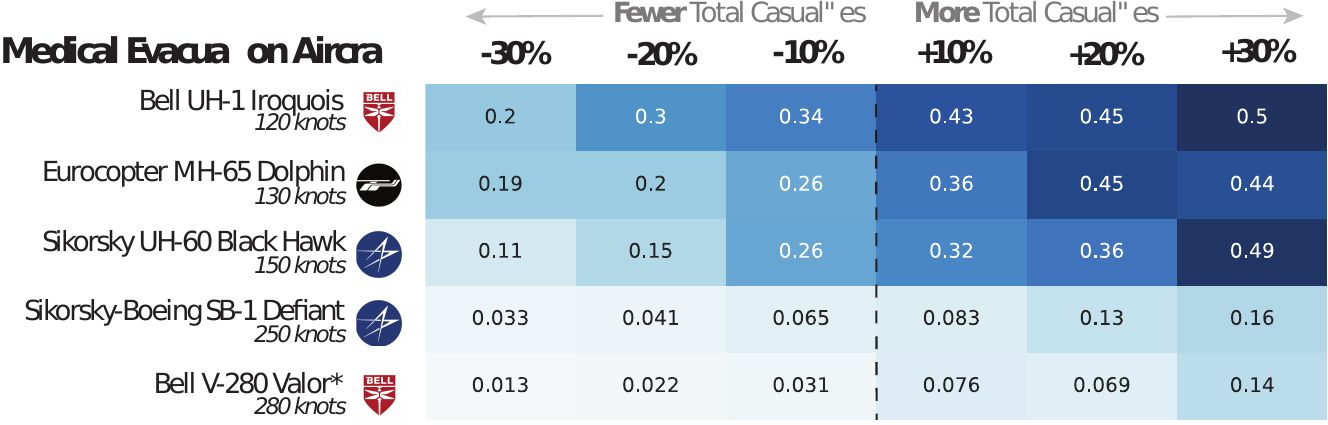}
\caption{Watercraft exchange point availability impact on incident response times across aircraft cruise speeds and casualty magnitudes. Five aircraft are considered, each with a historic, current, or proposed role in medical evacuation.}
\vspace{-1em}
\end{figure}

\section{Deployment}
We move from simulation to real-world deployment by facilitating a patient transfer with a manikin in Hawaii on October 2023 using two HH-60M Black Hawk helicopters and Army LSV-3 \textit{General B. Somervell} as a watercraft exchange point. Both HH-60M were launched from Wheeler Army Airfield, located in central Oahu. The mock patient was picked up from the ground force three miles east of Wheeler Army Airfield, transported to and from LSV-3 via aircraft as shown in Fig. 6, and ultimately delivered to Tripler Army Hospital. An execution checklist with aircraft dispatch times, informed by our decision process and solver, was provided to command posts located on Wheeler Army Airfield and LSV-3. Fig. 7 depicts the real-world inputs to our decision process and solver. These include the number of transportable patients in the 9-line evacuation request, participating evacuation platform characteristics, and real-time updates on platform locations and delays induced by non-participating air and maritime traffic. LSV-3 was postured ten miles south of Honolulu and moving in a south-westerly direction at five knots during patient drop-off and pickup. 

Fig. 8 compares real-world deployment outcomes to the simulation results which informed the real-world aircraft dispatch times. Boxes one and three annotate aircraft airfield departures. Box two annotates first aircraft arrival at the watercraft. Box four annotates the patient hand-off; only three minutes separated the first aircraft departing LSV-3 from the second aircraft arriving at LSV-3. The incident response time from evacuation request notification to role three patient delivery was 76 minutes, 16 minutes of which were caused by maritime traffic delays, annotated by the asterisk box with bars. Box six annotates the first aircraft's return to the airfield, box five annotates the second aircraft's departure from the watercraft, and box seven annotates the second aircraft's arrival at the role three. To maintain schedule, the second aircraft's airfield departure was automatically delayed commensurate to maritime traffic delays experienced by the first aircraft. This built-in update infrastructure enables continuous evacuation operations despite the non-stationary and unpredictable nature of the operating environment. 

This deployment resulted in the watercraft exchange point capability being recommended for addition to Army doctrinal publication ATP 4-02.2, \textit{Medical Evacuation}. This establishes the watercraft exchange point as a standard technique for 300-plus Army medical evacuation helicopters across more than 25 air ambulance companies and detachments.

\begin{figure}[ht]
    \centering
    \subfloat[\centering ][$\copyright$ Tristan Moore 2023]{{\includegraphics[width=0.519 \columnwidth]{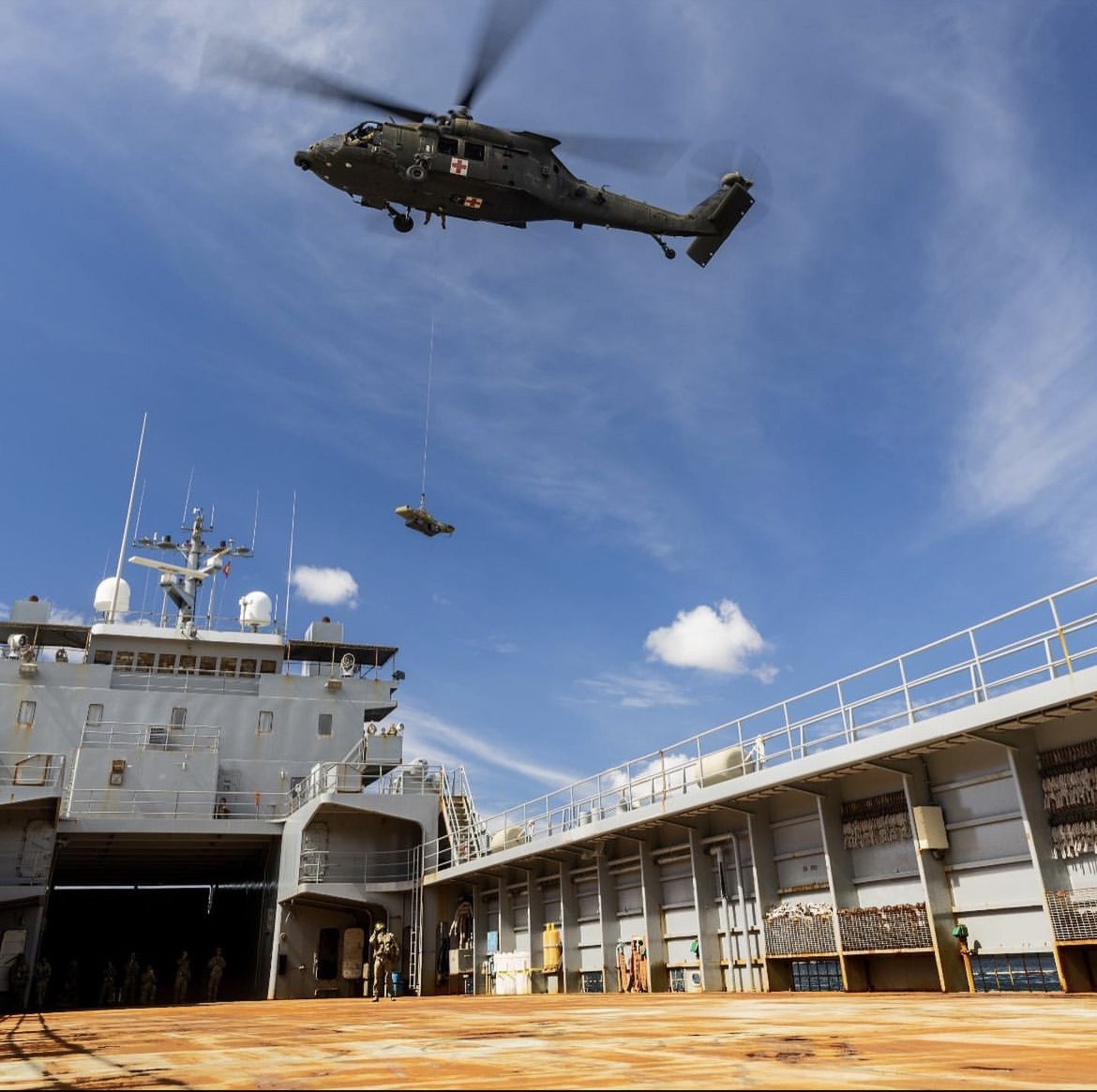} }}%
    \quad
    \subfloat[\centering ][$\copyright$ Tristan Moore 2023]{{\includegraphics[width=0.415 \columnwidth]{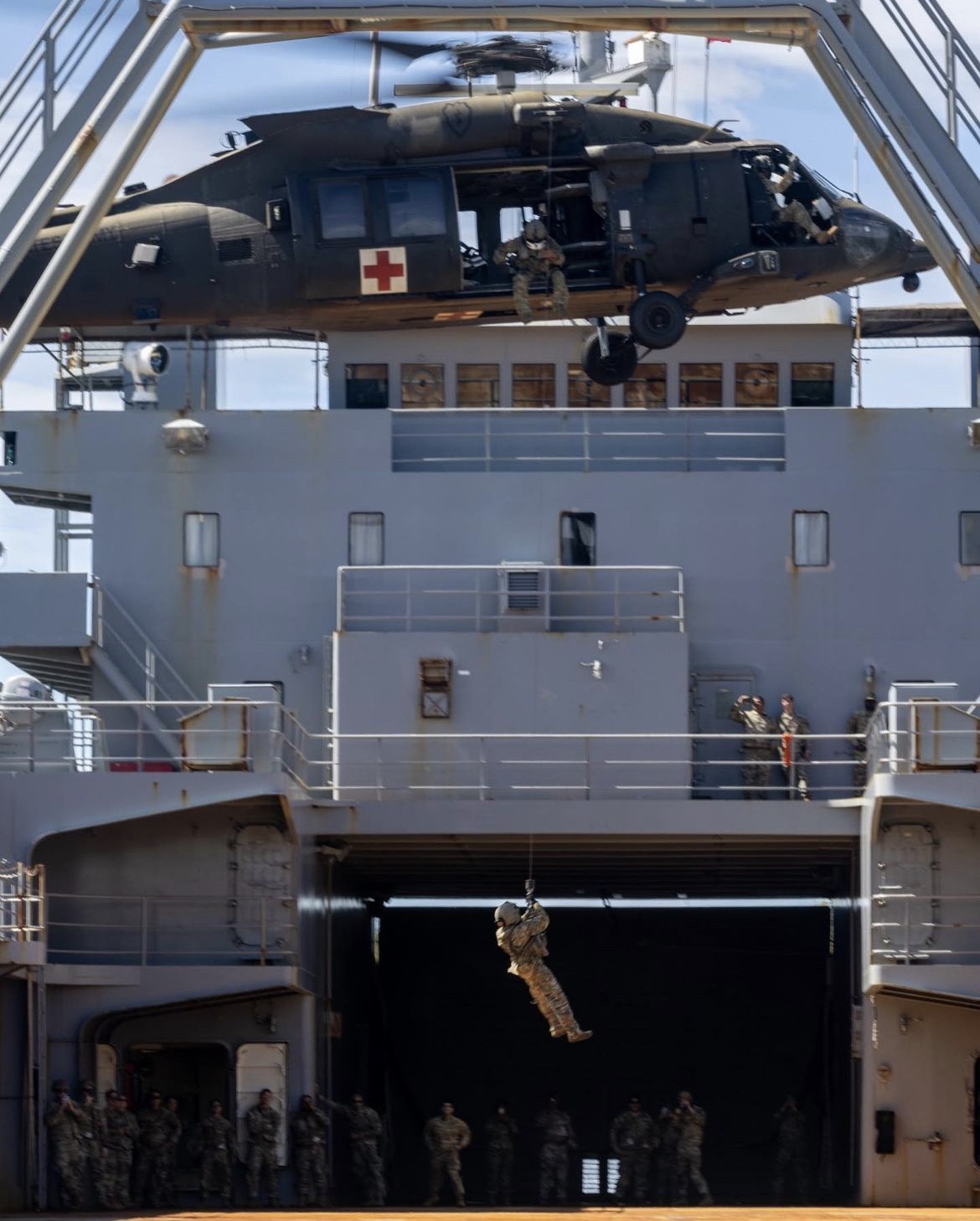} }}%
    \caption{An HH-60M Medical Evacuation Black Hawk helicopter executes hoist iterations to Army Logistics Support Vessel 3, which is traversing the open ocean south of Honolulu, Hawaii at 5 knots. These operations were  part of a two-week patient transfer exercise demonstrating the use of Army watercraft as ambulance exchange points.}%
    \vspace{-1em}
\end{figure}

\begin{figure*}[ht!]
\centering
\includegraphics[width=1\textwidth]{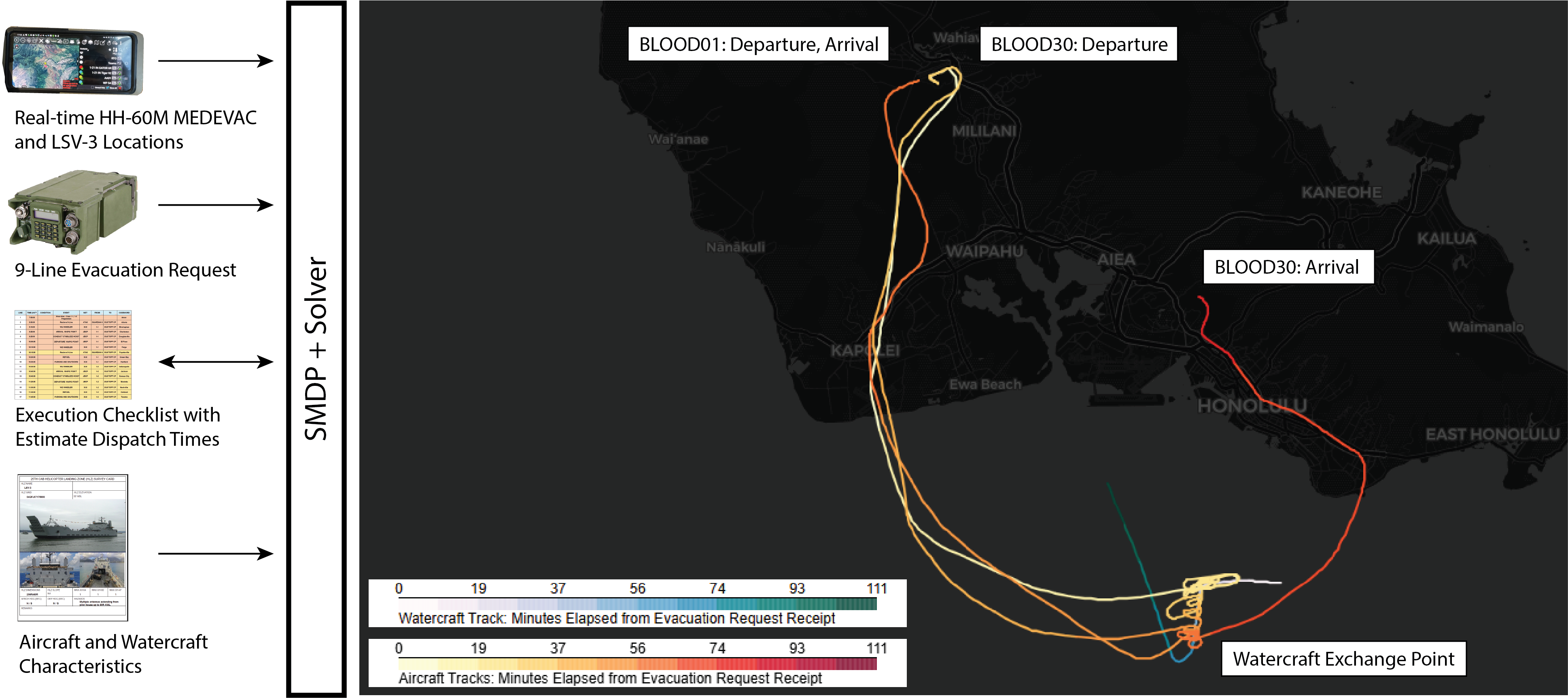}
\caption{Schematic architecture of the decision process and solver, and resulting ground track of two HH-60M Black Hawks (call signs BLOOD01 and BLOOD30) and the participating Logistics Support Vessel during deployment on 11 October 2023.}
\end{figure*}

\begin{figure}[ht!]
\centering
\includegraphics[width=1.0\columnwidth]{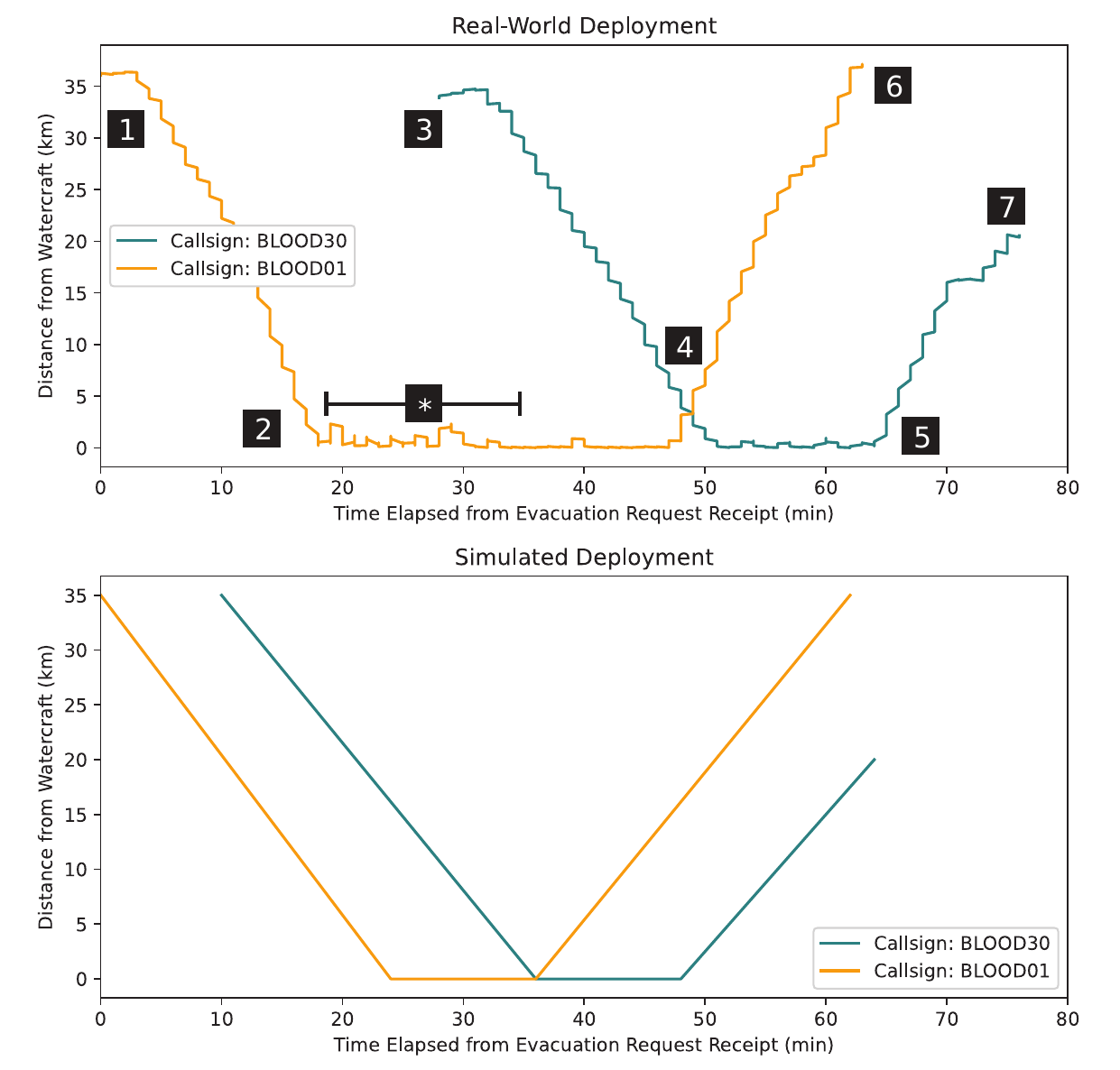}
\caption{Annotated comparison of deployment and simulation evacuation times and distances from the underway watercraft exchange point. Key insight: the simulation model accurately captures the real-world dynamics.}
\vspace{-1em}
\end{figure}

\section{Discussion}

An optimal policy with watercraft exchange points significantly outperforms an optimal policy without watercraft exchange points across specified ranges for all five examined casualty flow parameters in terms of both total rewards and incident response times. The presence of watercraft exchange points generally results in a 30-35\% improvement in total rewards. Similarly, an optimal policy without watercraft exchange points significantly outperforms a greedy policy across specified ranges for three of the five examined parameters, demonstrating a more modest 10-15\% improvement. We find a trade-off between reducing FSMP response times and increasing ASMP response times, such that the time reduced for the FSMP greatly outweighs the time increased for the ASMP. This imbalance indicates a geographically derived inefficiency in the dynamic resource allocation problem. Different geographies and operational scenarios will result in unique demand imbalances that may fluctuate over the course of a battle. Watercraft exchange points then play a balancing role, such that the time required to service a particular evacuation request can be partitioned to best support each island's unique utilization. 

Impressively, in nearly all examined instances, an optimal policy with watercraft exchange points minimizes the disparity in incident response time between platoons to near zero. For this to occur without any say over the location and movement of participating watercraft suggests that watercraft density may be a sufficient substitute for watercraft control. Without watercraft exchange points, even an optimal policy cannot bridge imbalances in demand, and the disparity in response time between platoons balloons up to hundreds of minutes for certain parameter configurations. Partitioning the transfer between ambulances comes at a cost: the patient handover process at any exchange point increases the total time to deliver a patient to their destination. This cost is often outweighed by the benefit induced by partitioning. 

Several insights were realized during deployment that support the future implementation of decision support tools for cooperative aerial and maritime evacuation planning. 

\begin{itemize}
  \item \textbf{Continuous Monitoring and Evaluation}: Evacuation operations often occur in high-density traffic areas that can induce substantial delays. Yet patient outcomes are tied to incident response times. Sensors for identifying and evaluating mission impact are therefore critical. 
  \item \textbf{Flexibility, Adaptability, and Risk}: The coordination of multiple platforms with asynchronous actions is susceptible to sudden changes in demand signal or environment. Update infrastructure must gauge risk when determining whether and how to amend existing guidance. 
  \item \textbf{Communication and Contingencies}: Assuming full observability requires platforms to communicate intentions and state information on a regular and predictable basis. Contingencies and partial observability models should be considered for when communication fails. 
\end{itemize}

\section{Conclusion}

Watercraft exchange points increase medical evacuation flexibility and reach in maritime environments. We demonstrate how watercraft exchange points enable patient transfers that correct for imbalances in aircraft utilization across islands. This results in significantly reduced incident response times. Although advantageous, watercraft exchange points are complicated by their non-dedicated and underway nature. Watercraft location and movement, and participating aircraft utilization, must be considered. We formulate an SMDP according to a relevant operational scenario and simulate using MCTS with root parallelization to solve the exchange point selection and aircraft dispatching problems. Platoon response times are determined across five parameters corresponding to casualty flow and aircraft characteristics. Integrating watercraft exchange points into evacuation planning results in a 35\% to 40\% improvement over existing methods. We deploy the watercraft exchange point model in the Hawaiian Islands using two HH-60M MEDEVAC helicopters and an Army LSV, which led to proposed revisions to Army doctrinal publications on medical evacuation.

\section{Acknowledgments}
This research was supported by the United States Army LSV-3 and LSV-7 watercraft crews and 545th Harbormasters. We are also grateful for C/3-25 Aviation Regiment ``Lightning DUSTOFF'' based out of Wheeler Army Airfield, Hawaii. Their HH-60M aircrews trained for months in the lead-up to the deployment, during which they executed the watercraft exchange point to great effect. 
\\~\\
All views expressed are those of the authors alone, and do not reflect the views of the US Army.

\clearpage

\bibliography{aaai25}

\end{document}